\documentclass[lettersize,journal]{IEEEtran}
\usepackage{amsmath,amsfonts}
\usepackage{algorithm}
\usepackage{array}
\usepackage[caption=false,font=normalsize,labelfont=sf,textfont=sf]{subfig}
\usepackage{textcomp}
\usepackage{booktabs}
\usepackage{stfloats}
\usepackage{url}
\usepackage{verbatim}
\usepackage{graphicx}
\usepackage{cite}
\usepackage{algorithm}
\usepackage{algpseudocode}
\usepackage{xcolor}
\usepackage{amssymb}
\usepackage[colorlinks,linkcolor=cyan,urlcolor=cyan]{hyperref}
\begin{document}

\title{FLoKD: Adaptive Knowledge Distillation for
Federated Low-Rank LLM over Wireless Networks}
\author{  Xinlu Zhang, Na Yan, Yang Su, Yansha Deng, \textit{Senior Member} and Toktam Mahmoodi, \textit{Senior Member}.
\thanks{X.~Zhang, N.~Yan, Y.~Su, Y.~Deng, and T.~Mahmoodi are with the Department of Engineering, King's College London, Strand, London WC2R 2LS, U.K. (e-mail: \{xinlu.zhang, na.yan, yang.su, yansha.deng, toktam.mahmoodi\}@kcl.ac.uk).

}}

\markboth{Journal of \LaTeX\ Class Files,~Vol.~14, No.~8, August~2021}%
{Shell \MakeLowercase{\textit{et al.}}: A Sample Article Using IEEEtran.cls for IEEE Journals}


\maketitle

\begin{abstract}
Large language models (LLMs) have demonstrated strong capabilities across a wide range of natural language processing tasks. However, conventional fine-tuning typically relies on centralized data collection, bringing in privacy concerns. Federated learning (FL) enables collaborative LLM fine-tuning without sharing raw client data, but its deployment over bandwidth-constrained wireless networks is hindered by the communication overhead of model-parameter transmission. Although Low-Rank Adaptation (LoRA) reduces the number of trainable parameters, its communication cost still increases with model scale. Knowledge distillation avoids parameter sharing via output logits, but token-level logits in LLMs incur high communication cost due to sequence length and vocabulary size. Reducing logits lowers the cost but weakens supervision and degrades accuracy. To address these limitations, we propose FLoKD, an adaptive knowledge-distillation framework for federated LoRA fine-tuning of LLMs over wireless networks, which communicates intermediate LoRA activations as the distillation signal rather than logits or full parameters. Since transmitting all blocks over the entire public dataset remains costly, we further propose a transformer block importance scoring framework that selectively transmits the most informative blocks, and two dataset selection strategies that discard public samples deviating from the local data distribution and prioritise those most informative for distillation. Extensive experiments across multiple generative language datasets, including  WikiText-103, PTB, and Dialog, demonstrate that our proposed framework reduces communication overhead by 50-65\% while achieving rapid convergence to competitive perplexity compared to baselines.
\end{abstract}


\section{Introduction}

\IEEEPARstart{R}{ecent} advances in pretrained large-scale language models, such as ChatGPT, LLaMA2, and BERT \cite{touvron2023llama, devlin2019bert}, have significantly improved natural language processing capabilities and accelerated the widespread adoption of LLMs in everyday applications.
To achieve better performance over diverse downstream tasks, even pretrained LLMs with strong generalization capabilities typically require additional fine-tuning on large amounts of data, often collected from multiple users or organizations. In conventional centralized fine-tuning, such data must be transferred to and stored on a central server, raising concerns on data privacy and security. Federated learning (FL) offers as an alternative paradigm, in which clients keep their data locally and exchange only model updates with the server, allowing multiple parties to collaboratively fine-tune a shared LLM without exposing their raw data\cite{li2023distributed}.

Although FL reduces the direct exposure of raw client data, its collaborative training requires multiple communication rounds between clients and the aggregation server. When applied to LLMs with billions of parameters, repeatedly exchanging model updates leads to substantial communication overhead, particularly over bandwidth-constrained wireless networks. For example, transmitting the full parameters of LLaMA3-8B in half precision requires approximately 16 GB per communication round. Meanwhile, full fine-tuning also imposes heavy client side computational and memory demands due to model parameters, gradients, and optimizer states. These communication and computation constraints make full model federated fine-tuning impractical for resource-constrained edge devices, motivating the use of parameter-efficient fine-tuning methods \cite{cheng2024toward,zhang2024joint,li2024federated}.

To alleviate the substantial communication overhead of transmitting full model parameters, parameter-efficient fine-tuning (PEFT) methods update only a small subset of parameters, reducing both local computation and communication costs while preserving downstream adaptation capability \cite{han2024parameter,zhang2026tt,zhang2025parameter,zhang2025communication}. Among them, Low-Rank Adaptation (LoRA) has become a widely adopted approach for federated LLM fine-tuning. LoRA freezes the pretrained model parameters and inserts trainable low-rank matrices into target modules, so that only lightweight adapters are updated and exchanged. This makes LoRA particularly suitable for federated LLM systems operating under wireless resource constraints.

Building upon this, federated LoRA fine-tuning methods have been proposed to enable collaborative LLM adaptation with significantly reduced communication costs by transmitting only compact LoRA parameters rather than full model weights \cite{bian2025survey,chai2021fedat}. However, existing federated LoRA methods still face two main limitations: the communication cost of LoRA transmission increases with model scale, and LoRA parameters are generally transmitted uniformly across transformer blocks without considering their varied contributions to knowledge transfer. First, the size of LoRA adapter matrices is directly determined by the hidden dimension of the blocks to which they are applied. As LLMs continue to scale up, their hidden dimensions also increase, producing larger LoRA matrices even when the adapter rank remains unchanged. Consequently, the communication cost of transmitting LoRA parameters grows with model size, potentially weakening the communication-efficiency advantage of full-LoRA transmission. Recent works attempt to alleviate this cost within the same paradigm, for example through adaptive quantisation under heterogeneous client bandwidth \cite{su2026haflq}, but the transmitted content remains the LoRA parameters themselves. Second, LoRA adapters are typically transmitted uniformly across transformer blocks, implicitly assuming that all blocks are equally important for knowledge transfer. Such an indiscriminate strategy may introduce unnecessary communication overhead from blocks that contribute little to global model alignment, while failing to prioritise more informative blocks. Existing federated LoRA methods therefore remain largely tied to parameter transmission, while adapting the granularity of the uplink signal according to block level importance has received little attention.

As an alternative to direct parameter exchange, knowledge distillation (KD) based federated learning has received increasing attention for communication-efficient knowledge transfer. FedMKT \cite{fan2025fedmkt} is a representative method in this direction, in which clients transmit output logits and subsequently distill the aggregated logits into their local models to enable mutual knowledge transfer. However, this logit-based design has several limitations. First, token-level logits scale with both sequence length and vocabulary size and may therefore still incur substantial communication overhead for LLMs. Second, output logits describe only the final prediction distribution and do not preserve the richer intermediate representations learned across transformer blocks, potentially limiting the amount of transferable knowledge. Third, FedMKT relies on a shared public dataset without explicitly evaluating the quality or task relevance of individual samples, making the distillation process sensitive to noisy or poorly aligned public data.

Recent studies have therefore explored intermediate representations as richer sources of distillation knowledge. LRC \cite{haotoken}, for example, proposes a low-rank clone framework that distills smaller language models from larger teachers by aligning intermediate activations rather than relying solely on output logits. Similarly, DDK \cite{liu2024ddk} incorporates domain-aware intermediate signals to dynamically guide the distillation process. These methods suggest that intermediate layer representations can provide more informative supervision than final output distributions alone. Nevertheless, they are primarily designed for centralized training and are not directly communication efficient in bandwidth constrained federated settings. Therefore, although intermediate representations provide richer knowledge for distillation, their practical use in federated LLM fine-tuning requires the selective communication of only the most informative samples and transformer blocks.

To address the limitations of previous work, we propose a novel \textbf{Adaptive Knowledge Distillation for Federated Low-Rank LLM over Wireless Networks} (FLoKD) framework, which compresses the uplink signal by transmitting intermediate LoRA activations instead of model parameters or output logits, and then reducing the transmitted activations along the block and the sample dimension through block importance scoring and distillation-oriented public data selection. Our contributions can be summarized as follows:
\begin{itemize}
    \item \textbf{Intermediate Activation-Based Knowledge Distillation for Federated LLM Fine-Tuning.} To overcome the limited internal representation information provided by output logits, we propose a novel knowledge distillation framework that transmits the intermediate activations of LoRA blocks instead of output logits. The dimension of these activations is far smaller than that of output logits, incurring significantly less communication cost. Unlike logits, which only capture the final compressed prediction distribution, intermediate activations retain rich internal representations learned within the transformer blocks, enabling more effective knowledge transfer across clients while maintaining low communication overhead.
  \item \textbf{Transformer Block Importance Evaluation for Block Selection.} Although a single LoRA activation is small in size, each transformer block produces its own activations. Transmitting all of them therefore incurs a communication cost that grows linearly with model depth. To reduce this overhead, we propose an importance scoring framework that quantitatively evaluates the contribution of each transformer block based on its intermediate activations. By selectively transmitting only the activations from the most important blocks while skipping less informative ones, our proposed framework significantly reduces the volume of transmitted data without sacrificing knowledge transfer quality.
    \item \textbf{Token-Aware and Loss-Based Public Data Selection.} Distillation over a public dataset that is misaligned with the clients' private data distributions transfers knowledge irrelevant to the downstream task, degrading distillation efficiency. To address this, we propose two complementary and efficient strategies for selecting the public dataset. Token-aware selection serves as a one-off filtering method that aligns public samples with the clients' token distribution, removing distributionally misaligned samples while substantially reducing computation and communication costs. In contrast, loss-based selection prioritizes the samples on which the model incurs high loss, retaining those most informative for knowledge transfer and achieving strong overall performance. Together, the two strategies improve both distillation efficiency and convergence stability under heterogeneous data.
    \item \textbf{Pareto-Optimal Communication Trade-Off Analysis.} We formulate the allocation of a fixed communication budget between the number of selectively transmitted Transformer blocks and the size of the selected distillation dataset as a constrained resource allocation problem. By deriving the corresponding bandwidth trade-off and empirically characterizing the Pareto frontier between block coverage and data coverage, we provide a systematic and practical guideline for jointly configuring our two proposed mechanisms under a given wireless bandwidth constraint.
    \item \textbf{Comprehensive Experimental Evaluation.} Our proposed framework is evaluated over three language modelling benchmarks, namely WikiText-103, Penn Treebank, and DailyDialog. Ablation studies further isolate the contribution of block importance scoring and sample selection, and the results show that FLoKD achieves competitive perplexity under tight uplink budgets where parameter-based and logit-based baselines degrade noticeably.
\end{itemize}
The rest of this paper is organized as follows: Section II introduces the system model. Section III presents the problem formulation, block importance scoring, public-data selection, communication analysis, and block-data allocation strategy. Section IV provides the numerical results and performance evaluation of the proposed framework. Finally, Section V concludes the paper.

\begin{figure*}[t!]
  \centering
  \includegraphics[width=1\textwidth]{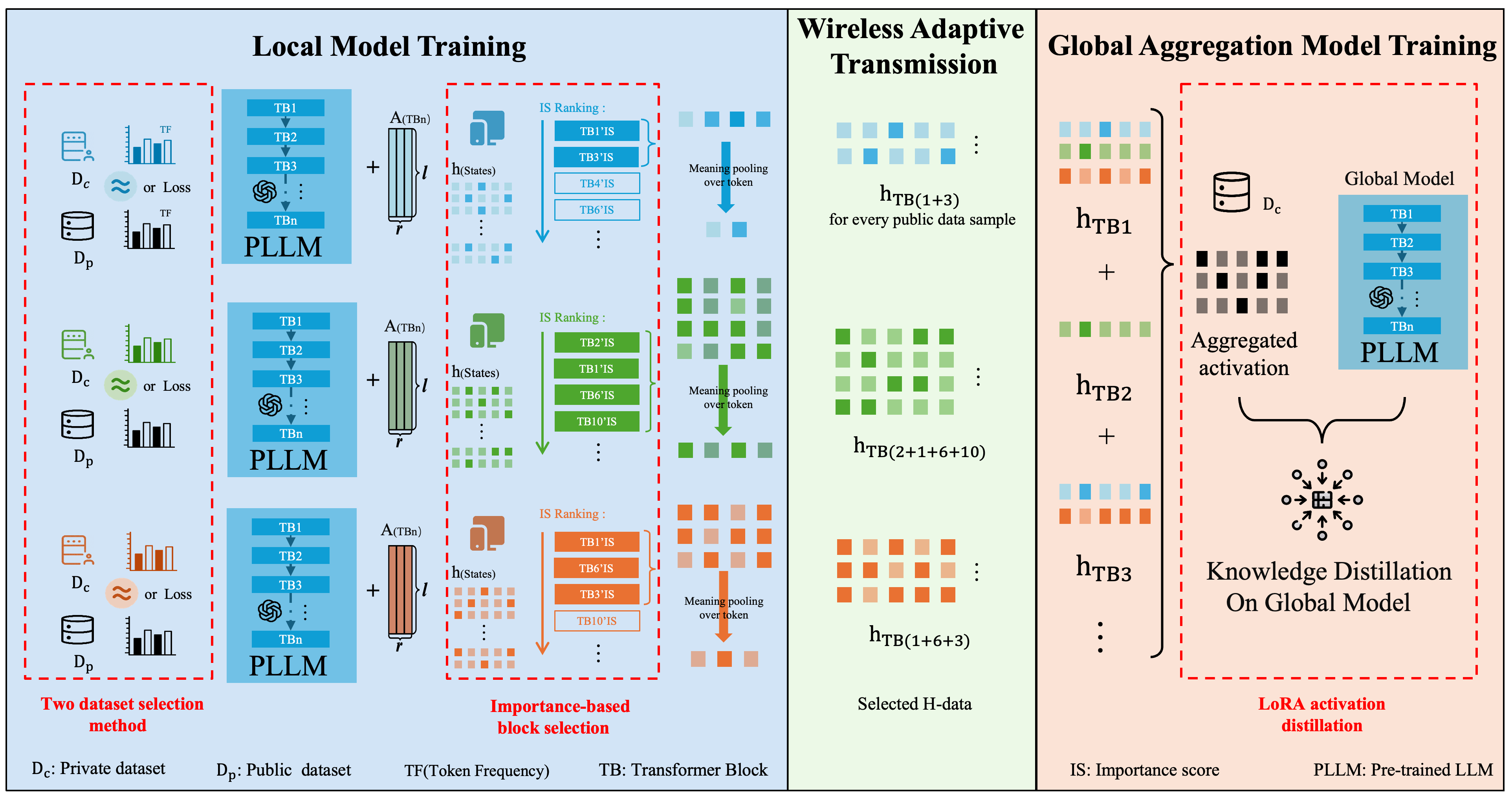}
  \caption{The workflow of the proposed FLoKD framework.  }
  \label{fig1}
\end{figure*}

\section{System Model}

Consider a Federated Learning network consisting of a cloud server $\mathcal{S}$ and $N$ distributed clients, denoted $\mathcal{N} = \{1, 2, \ldots, N\}$. The server and the clients hold LLMs fine-tuned by LoRA with parameters $\theta_g$ and $\theta_n$ under the same backbone architecture with shared weights $W'$, where $W'$ denotes the frozen pre-trained backbone weights shared across all parties, while $\theta_n$ represents the client-specific trainable LoRA parameters. The server and clients share the same public dataset $\mathcal{D}_p$, with a sample volume $|\mathcal{D}_p|$. Each client $n \in \mathcal{N}$ trains on a private generative dataset $\mathcal{D}_n$, with a data volume $|\mathcal{D}_n|$. Each data sample is represented as a token sequence $\mathbf{X}_{n,i}$, $i \in \{1, 2, \ldots, |\mathcal{D}_n|\}$, where $\mathbf{X}_{n,i}$ denotes the $i$-th input token sequence, truncated or padded to a maximum length $T_max$. The local datasets $\{\mathcal{D}_n\}_{n=1}^{N}$ are non-independent and non-identically distributed (non-IID) across clients, exhibiting heterogeneous token distributions.

The server and clients collaboratively improve the global model through federated knowledge distillation by exchanging LoRA activations computed on public data, rather than model parameters. As shown in Fig.~\ref{fig1}, each communication round consists of the following steps.

\subsection{Local Fine-Tuning and Upload}
In our framework, a pretrained model is collaboratively fine-tuned across distributed clients without exchanging their raw private data. Due to the large scale of LLMs and the limited resources available at clients, we employ LoRA for parameter-efficient local adaptation\cite{hu2024federated,cheng2024toward}. Motivated by the observation that task-specific weight updates lie approximately in a low-dimensional subspace \cite{hu2022LoRA}, LoRA updates only a small set of low-rank matrices while freezing the pretrained backbone.

Specifically, consider a linear layer with a pretrained weight matrix $W' \in \mathbb{R}^{d_{\mathrm{out}} \times d_{\mathrm{in}}}$, which is shared by all clients and kept frozen throughout training, where $d_{\mathrm{in}}$ and $d_{\mathrm{out}}$ denote the input and output dimensions of the layer, respectively. Instead of directly updating the full weight matrix, each client $n$ introduces a trainable low-rank update parameterised by $\theta_n=\{A_n,B_n\}$ and updated on its own private dataset. The adapted weight matrix is defined as
\begin{equation}
\setlength{\belowdisplayshortskip}{2pt}
\label{eq:LoRA}
W_n = \underbrace{W'}_{\text{shared, frozen weight}} + \underbrace{B_n A_n}_{\text{trainable low-rank update}},
\end{equation}
where $A_n \in \mathbb{R}^{r \times d_{\mathrm{in}}}$ projects the input features into an $r$-dimensional low-rank subspace, and $B_n \in \mathbb{R}^{d_{\mathrm{out}} \times r}$ maps the resulting low-rank representation back to the original output space. The rank $r$ is much smaller than both the input and output dimensions, i.e., $r \ll \min(d_{\mathrm{in}},d_{\mathrm{out}})$ \cite{zhang2025communication}. 

During local training, each client keeps $W'$ fixed and updates $A_n$ and $B_n$ via empirical risk minimisation over its local dataset. Accordingly, the objective of each client is formulated as
\begin{equation}
\small
\begin{split}
\mathcal{L}_n(\theta_n) ={}& -\frac{1}{\lvert \mathcal{D}_{n}\rvert} \sum_{\mathbf{X}_{n,i}\in \mathcal{D}_{n}} \frac{1}{T_{n,i}}\sum_{t=1}^{T_{n,i}} \\
& \log p\bigl(x_{n,i}^{t} \mid x_{n,i}^{<t};\,W',\theta_n\bigr) +\lambda\,\mathcal{R}(\theta_n)\,,
\end{split}
\label{eq1}
\end{equation}
where $T_{n,i}$ denotes the length of sequence $\mathbf{X}_{n,i}$, the first term is the average per-token negative log-likelihood that measures how well the model predicts each token $x_{n,i}^{t}$ given its preceding context $x_{n,i}^{<t}$ over all sequences in $\mathcal{D}_n$, and $\mathcal{R}(\theta_n)$ is an optional regularizer on $\theta_n$ with weight $\lambda$ that suppresses overfitting to the limited local data. The minimization is performed over several local iterations before the client interacts with the server.

\subsection{Sample selection and Uplink Transmission}
After local fine-tuning, each client $n$ transfers its locally learned knowledge to the server by uploading intermediate LoRA activations computed on the public dataset, rather than the LoRA parameters exchanged in conventional federated fine-tuning frameworks. To satisfy the uplink bandwidth constraint, client $n$ applies the data- and block-selection strategies introduced in Section~III, resulting in $P$ selected public samples and a client-specific set $\mathcal{K}_n$ of selected Transformer blocks, with $|\mathcal{K}_n|=K$. For the $i$-th selected sample and a selected block $k\in\mathcal{K}_n$, let $\mathbf{x}^{(k,i)}$ denote the input activation to block $k$.

For an input activation $\mathbf{x}^{(k,i)}$ at block $k \in \mathcal{K}_n$ and selected public sample $i$, the intermediate LoRA activation is given by
\begin{equation}
\mathbf{h}_{n}^{(k,i)} = {A}_{n}^{(k)} {x}^{(k,i)}
\in \mathbb{R}^{r},
\label{eq7}
\end{equation}
where ${A}_{n}^{(k)}$ is the LoRA down-projection matrix of client $n$ at block $k$. Stacking these vectors over the $P$ selected samples gives the block-wise activation matrix
\begin{equation}
\mathbf{H}_{n}^{(k)}
=
\begin{bmatrix}
(\mathbf{h}_{n}^{(k,1)})^\top \\
\vdots \\
(\mathbf{h}_{n}^{(k,P)})^\top
\end{bmatrix}
\in \mathbb{R}^{P \times r}.
\label{eq:block_activation_matrix}
\end{equation}
Collecting the block-wise matrices over all selected blocks yields the uploaded activation tensor
\begin{equation}
\mathbf{H}_{n}
=
\bigl[\mathbf{H}_{n}^{(k)}\bigr]_{k \in \mathcal{K}_n}
\in \mathbb{R}^{P \times K \times r},
\label{eq:activation_tensor}
\end{equation}
where the matrices $\mathbf{H}_{n}^{(k)}$ are stacked along the block dimension. In this way, only the most informative LoRA activations are transmitted to the server.

\subsection{Global Aggregation and Knowledge Distillation}

After receiving the selected LoRA intermediate activations from all $N$ clients, the server aggregates them for the subsequent distillation. Since different clients may select different transformer blocks, the uploaded activations cannot be directly averaged. The server instead aligns them by block index and aggregates within each block separately.

A naive treatment is to average over all $N$ clients regardless of participation, implicitly treating the missing activations of non-contributing clients as zeros. This scales down the aggregated activation of any block selected by only a subset of clients, distorting its magnitude. To avoid such dilution, the server aligns the uploaded activations by block index and aggregates only over the contributing clients within each block. Let $\mathcal{K}^{*} = \bigcup_{n=1}^{N} \mathcal{K}_n$ denote the set of blocks selected by at least one client. For each block $k \in \mathcal{K}_{*}$, let $\mathcal{N}_k$ denote the set of clients that have selected block $k$. The aggregated activation of block $k$ is then given by
\begin{equation}
\mathbf{H}_{\mathrm{a}}^{(k)} = \sum_{n \in \mathcal{N}_k} \alpha_n^{(k)}\, \mathbf{H}_n^{(k)},
\qquad
\alpha_n^{(k)} = \frac{1}{|\mathcal{N}_k|},
\label{eq:aggregation}
\end{equation}
where $\alpha_n^{(k)}$ is the aggregation weight of client $n$, set uniformly over the contributing clients in $\mathcal{N}_k$. The aggregated client knowledge is the block-indexed set $\{\mathbf{H}_{\mathrm{a}}^{(k)}\}_{k \in \mathcal{K}_{*}}$.

To transfer the aggregated client knowledge into the global model $W_{\mathrm{g}}$, the server aligns the global model's intermediate LoRA activations with the aggregated client activations on the selected public dataset $\mathcal{D}_{\mathrm{p}}^{*}$. Let $\mathbf{h}_{\mathrm{g}}^{(k,i)} = A_{\mathrm{g}}^{(k)}\mathbf{x}_{\mathrm{g}}^{(k,i)} \in \mathbb{R}^{r}$ denote the intermediate LoRA activation of the global model at block $k$ for the $i$-th selected public sample, obtained in the same manner as Eq.~\eqref{eq7} with $ A_{\mathrm{g}}^{(k)}$ the LoRA down-projection matrix of $W_{\mathrm{g}}$, and let $\mathbf{h}_{\mathrm{a}}^{(k,i)} \in \mathbb{R}^{r}$ denote the $i$-th row of $\mathbf{H}_{\mathrm{a}}^{(k)}$.

In conventional knowledge distillation, knowledge is commonly transferred through output logits, and the Kullback--Leibler (KL) divergence is used to match the softened output distributions of the teacher and student models~\cite{hinton2015distilling,chen2024trakdis}. In our framework, however, the distilled quantities are intermediate LoRA activations, which are continuous representation vectors rather than probability distributions. KL divergence is therefore not directly applicable to our proposed activation-based knowledge transfer. Instead, we employ cosine distance to measure the directional discrepancy between the global-model and aggregated client activations~\cite{xia2015learning}\cite{liu2026crb}. This metric is particularly suitable for our setting for two reasons. First, cosine distance is invariant to positive rescaling of the activation vectors, allowing the distillation objective to focus on their representational directions rather than their absolute magnitudes. This is important because heterogeneous local data and training dynamics may result in scale differences between the aggregated client activations and the corresponding global-model activations, which could otherwise dominate a direct value-matching objective. Second, cosine distance is bounded, providing a stable and consistently scaled objective compared with an unbounded squared-error loss that may increase substantially with activation magnitude. Minimising the resulting objective therefore aligns the global activations with the aggregated client activations.

Specifically, the sample-wise cosine distance is defined as
\begin{equation}
d_{\mathrm{cos}}(\mathbf{a},\mathbf{b})
=
1-
\frac{
\langle \mathbf{a},\mathbf{b} \rangle
}{
\|\mathbf{a}\|_{2}\|\mathbf{b}\|_{2}+\epsilon
},
\label{eq:cosine_distance}
\end{equation}
where $\epsilon>0$ is a small constant used for numerical stability. The server-side distillation loss is then defined as
\begin{equation}
\mathcal{L}_{\mathrm{server}}
=
\frac{1}{P|\mathcal{K}_{*}|}
\sum_{k\in\mathcal{K}_{*}}
\sum_{i=1}^{P}
d_{\mathrm{cos}}
\left(
\mathbf{h}_{\mathrm{a}}^{(k,i)},
\mathbf{h}_{\mathrm{g}}^{(k,i)}
\right).
\label{eq:server_loss}
\end{equation}

By computing the cosine distance independently for each sample and subsequently averaging over all selected samples and transformer blocks, our proposed formulation assigns equal weight to each selected public sample in the distillation objective. This avoids the implicit emphasis on samples with larger activation norms that may arise when cosine similarity is computed over the entire $P\times r$ activation matrix.

The server distils the aggregated client knowledge into the global model by minimising $\mathcal{L}_{\mathrm{server}}$. During this process, only the global LoRA parameters are updated, while the pretrained backbone remains frozen. At communication round $\tau$, the global LoRA matrices are updated as
\begin{equation}
\begin{aligned}
A_{\mathrm{g}}^{(k,\tau+1)}
&=
A_{\mathrm{g}}^{(k,\tau)}
-
\eta
\nabla_{A_{\mathrm{g}}^{(k)}}
\mathcal{L}_{\mathrm{server}},
\\
B_{\mathrm{g}}^{(k,\tau+1)}
&=
B_{\mathrm{g}}^{(k,\tau)}
-
\eta
\nabla_{B_{\mathrm{g}}^{(k)}}
\mathcal{L}_{\mathrm{server}},
\qquad k \in \mathcal{K}_{*},
\end{aligned}
\label{eq:LoRA_update}
\end{equation}
where $\eta$ is the learning rate, and the LoRA parameters of the unselected blocks remain unchanged in the current round. Although $\mathcal{L}_{\mathrm{server}} $ is defined directly on the down-projection outputs, the up-projection matrices $B_{\mathrm{g}}^{(k)}$ can also receive gradients indirectly through their influence on the input activations of downstream selected blocks.

After the server-side distillation is completed, the server broadcasts the updated global LoRA matrices, together with the indices of the selected public samples, to all clients. This ensures that every client uses the same public samples when computing its local activations in the subsequent communication round.

\subsection{Local Update on Clients}
After server-side distillation, the updated global LoRA matrices $\{\mathbf{A}_{\mathrm{g}}^{(k)}, \mathbf{B}_{\mathrm{g}}^{(k)}\}_{k\in\mathcal{K}^{*}}$ are broadcast to all clients, which replace the corresponding local LoRA parameters while the parameters of the remaining blocks are kept unchanged. Since the downlink is typically less constrained than the uplink in the considered wireless setting, transmitting the LoRA parameters directly is affordable and requires no additional local optimisation. Each client then resumes local fine-tuning on its private dataset $\mathcal{D}_{n}$ from the updated parameters, which restores the personalisation that the global parameters do not carry, and the next communication round proceeds as described above.

\section{Transformer Block Importance Scoring and Dataset Selection}

In this section, we present FloKD. We first formalize the token-level LoRA activations and their pooled representation used for transmission, and then propose two selection mechanisms based on our formulated problem.




\subsection{Problem Formulation}
\label{subsec:problem}

We consider a federated fine-tuning system in which a central server and $N$ clients collaboratively adapt a frozen backbone $W'$ via private low-rank parameters $\theta_n = \{\mathbf{A}_n, \mathbf{B}_n\}$ under non-IID partitions, with a server-held public dataset $\mathcal{D}_p$ as the medium for knowledge exchange.

Departing from parameter- or logit- based federation, our framework transfers knowledge through the intermediate LoRA activations evaluated on the public data. Each public sample $i$ is tokenized and truncated to a maximum length $T_{\max}$, producing $T_i \le T_{\max}$ tokens. Within block $k$ of client $n$, the LoRA update $\Delta\mathbf{W}_n^{(k)} = \mathbf{B}_n^{(k)}\mathbf{A}_n^{(k)}$ generates an intermediate activation through the low-rank down-projection,
\begin{equation}
    \mathbf{h}^{(i)}_{n,k,t} = \mathbf{A}_n^{(k)}\,\mathbf{x}^{(i)}_{n,k,t} \in \mathbb{R}^{r},
    \label{eq:activation}
\end{equation}
\noindent where $\mathbf{x}^{(i)}_{n,k,t}$ is the input to the adapted projection for the $t$-th token.


Since this per-token representation has size $T_{\max} \times r$ per sample and per block, transmitting it in full would dominate the uplink payload. Mean pooling is therefore applied over the valid token positions to obtain a single $r$-dimensional vector per sample and per block. We collapse the sequence dimension by mean pooling. It is presented as
\begin{equation}
    \bar{\mathbf{h}}^{(i)}_{n,k} = \frac{1}{T_i}\sum_{t=1}^{T_i} \mathbf{h}^{(i)}_{n,k,t} \in \mathbb{R}^{r},
    \label{eq:pooling}
\end{equation}
\noindent which yields a single $r$-dimensional descriptor per block, independent of $T_i$, and reduces the per-sample, per-block payload from $T_i r$ to $r$. Stacking these vectors over all public samples and blocks, the activations of client $n$ form a tensor $\mathbf{H}_n \in \mathbb{R}^{P \times L \times r}$, where $P$ is the number of public samples retained for transmission and\textit{ L} is the total number of transformer blocks. Formally, this reduces to minimising the cosine distillation loss in~\eqref{eq:server_loss}, which sums over the transmitted blocks $\mathcal{K}_{n}$ and averages over the $P$ public samples.



The server therefore acquires task-relevant knowledge without accessing any private data. Under wireless deployment, however, the uplink bandwidth available to each client is limited, so only a subset of these activations can be transmitted in each round.

In each communication round, client $n$ transmits the pooled activations of its $K = |\mathcal{K}_{n}|$ selected blocks over the selected public subset $\mathcal{D}^{*}_{p}$, where $|\mathcal{D}^{*}_{p}| = P$. This incurs an uplink cost formulated as 

\begin{equation} 
C_n = \rho \cdot P \cdot K \cdot r, 
\label{eq:comm_cost}
\end{equation}

\noindent where $\rho$ is the number of bits used to encode each activation element. Transmitting the indices of the selected public samples would introduce an additional $O(P)$ overhead, which is negligible compared with the activation payload and is therefore omitted. 

Constraining the per-client uplink to a budget $B$, the communication-aware fine-tuning problem can be formulated as

\begin{equation}
\begin{aligned}
\min_{\theta_g,\, \mathcal{K}_{n},\, \mathcal{D}^{*}_{p}} \quad 
& \mathcal{L}_{\text{server}}(\theta_g) \\
\text{s.t.} \quad 
& \rho \cdot P \cdot K \cdot r \leq B, \\
& \mathcal{K}_{n} \subseteq \{1, \dots, L\}, \\
& |\mathcal{K}_{n}| = K, \\
& \mathcal{D}^{*}_{p} \subseteq \mathcal{D}_{p}, \\
& |\mathcal{D}^{*}_{p}| = P.
\end{aligned}
\label{eq:constrained_problem}
\end{equation}


Problem~\eqref{eq:constrained_problem} separates two aspects of the transmitted content. The cardinalities $K$ and $P$ determine the communication cost and are jointly constrained by the uplink budget, whereas the choice of which blocks to include in $\mathcal{K}_{n}$ and which samples to retain in $\mathcal{D}^{*}_{p}$ determines how much useful knowledge the transmitted activations carry. Under a fixed budget, the objective is therefore to make the most informative selection at the given cardinalities, which motivates the block importance scoring and public data selection mechanisms developed below.

The remainder of this section develops a systematic answer to each, through block importance scoring (Section~\ref{subsec:block}) and public data selection (Section~\ref{subsec:data_selection}) respectively, and finally analyses how the budget should be split between the two.

Motivated by the goal of transmitting only the content that contributes to distillation\cite{shi2023task}, we address these two degrees of freedom with two complementary mechanisms. A \textbf{block importance scoring} strategy restricts activation uploads to the most informative Transformer blocks, thereby limiting the dependence of the payload on model depth. In parallel, two \textbf{public data selection} strategies select the public dataset through either token-aware distribution matching or loss-based ranking, so that activations are computed and transmitted only on the samples that contribute most to distillation. Since the per-round payload scales with the product of the two factors rather than with either alone, reducing both jointly is necessary to operate under a constrained uplink budget.


\subsection{Block Importance Scoring}
\label{subsec:block}
 
The backbone is composed of $L$ stacked Transformer blocks, each augmented with
its own LoRA adapter, so the most direct upload strategy would transmit the
intermediate activations of all $L$ blocks. However, transmitting all activations indiscriminately wastes communication resources and can even degrade performance. On the one hand, not all blocks
contribute equally to the adaptation objective: Many blocks respond only weakly to the fine-tuning signal and consequently contribute little task-relevant information. Uploading all block activations therefore consumes communication resources on redundant representations without providing meaningful gains in model adaptation.
 On the other hand,
the activations of these weakly informative blocks act as noise during
aggregation, diluting the signal contributed by the informative blocks and
degrading the quality of the distilled global model\cite{nisbett1981dilution},\cite{askari2026layerif}. 


We therefore adopt a selective upload strategy that transmits only the
activations of blocks that are both loss-sensitive and sufficiently adapted to
the current task. Since the pretrained weights stay frozen and the low-rank
update is zero-initialised, that update constitutes the entire deviation of a
block from the pretrained backbone, so its magnitude measures how strongly the
block has adapted to the local data.  Under this strategy, rather than uploading activations from all $L$ Transformer blocks, we estimate the importance of each block using a score that combines the loss sensitivity of its LoRA activation with the magnitude of its LoRA update. Blocks with high scores are considered more
informative for distillation. We then rank all blocks by this score and upload
only the Top-$K$ activations, discarding the remaining blocks to reduce
communication overhead and suppress redundant or weakly informative signals.

\paragraph{Importance Score}
For each block $k \in \{1, \dots, L\}$, we assign an importance
score to quantify the contribution of its LoRA-induced intermediate activation
to the training objective. The score follows from a first-order Taylor approximation of the loss\cite{molchanov2019importance}: a block is important when removing its LoRA-branch activation on the public calibration samples would change the loss appreciably, and this change is approximated to first order by the inner product between the loss gradient and the activation.

Therefore, we define a unified importance score $\mathrm{Score}(k)$ for each block as

\begin{equation}
    \mathrm{Score}(k) =
    \frac{1}{P}\sum_{i=1}^{P}
    \left|
    \left\langle
    \nabla_{\bar{\mathbf{h}}^{(i)}_{n,k}}\ell^{(i)},
    \bar{\mathbf{h}}^{(i)}_{n,k}
    \right\rangle
    \right|,
    \label{eq:block_score}
\end{equation}

\noindent where $\bar{\mathbf{h}}^{(i)}_{n,k} \in \mathbb{R}^{r}$ denotes the pooled
intermediate activation generated by the LoRA branch of block $k$ for the
$i$-th public calibration sample, corresponding to the $(i,k)$ slice of
$\mathbf{H}_n$; $\ell^{(i)}$ is the per-sample training loss evaluated on that
sample, and $P$ is the number of public calibration samples.

A block with a small estimated loss change is considered less important for the current objective and can therefore be deprioritized for transmission. The inner product
$\left\langle \nabla_{\bar{\mathbf{h}}^{(i)}_{n,k}}\ell^{(i)}, \bar{\mathbf{h}}^{(i)}_{n,k} \right\rangle$
measures, to first order, the change in the loss caused by removing the activation. Specifically, zeroing $\bar{\mathbf{h}}^{(i)}_{n,k}$ suppresses the low-rank contribution of block $k$ to its output, and the inner product represents the resulting first-order change in the loss. It thus reflects both how sensitive the loss is to the LoRA branch and how large that branch's activation is on the calibration data. A block whose removal would barely move the loss carries little task-relevant signal and can be dropped, whereas one whose removal would move it appreciably is retained. Averaging the magnitude of this quantity over the $P$ samples yields the score, following the first-order importance criterion used in network pruning~\cite{molchanov2016pruning,lee2018snip,}.

We use the inner product rather than the product of the gradient and activation
norms because the first-order loss change depends on their alignment. Writing
\[ \langle \nabla_{\bar{\mathbf{h}}}\ell,\, \bar{\mathbf{h}}\rangle = \|\nabla_{\bar{\mathbf{h}}}\ell\|_2\,\|\bar{\mathbf{h}}\|_2\,\cos\theta, \]
where $\theta$ is the angle between the two, a block with a large gradient and a
large activation still contributes little when they are nearly orthogonal. The
product of norms discards this factor and would over-rank such a block.

Although our activation is the low-rank bottleneck
$\bar{\mathbf{h}}^{(i)}_{n,k}=\mathbf{A}_k\mathbf{x}^{(i)}$, which does not depend on
$\mathbf{B}_k$, the score needs no separate term for the magnitude of the update
at block $k$. The dependence on $\mathbf{B}_k$ enters through the gradient: with
$\mathbf{g}_{\mathrm{out}}$ the gradient of the loss with respect to the LoRA
branch output, $\nabla_{\bar{\mathbf{h}}}\ell = \mathbf{B}_k^{\top}
\mathbf{g}_{\mathrm{out}}$, so that $\langle \nabla_{\bar{\mathbf{h}}}\ell,\,
\bar{\mathbf{h}}\rangle = \mathbf{g}_{\mathrm{out}}^{\top}\mathbf{B}_k\mathbf{A}_k
\mathbf{x}^{(i)}$ is already linear in $\mathbf{B}_k$ (and in $\mathbf{A}_k$).  The
score is therefore homogeneous of degree one in $\mathbf{B}_k$: rescaling the
adapter by a factor $c$ rescales the score by $|c|$, so the update magnitude is
already accounted for, though only in combination with
$\mathbf{g}_{\mathrm{out}}$, $\mathbf{A}_k\mathbf{x}^{(i)}$, and their alignment. Multiplying it by a norm such as $\|\mathbf{B}_k\|_F$ or
$\|\mathbf{B}_k\mathbf{A}_k\|_F$ would make it quadratic in the update magnitude,
over-ranking blocks whose adapter is merely large rather than relevant to the
current objective. Since $\mathbf{B}_k$ is zero-initialised, a block whose adapter
has scarcely moved already yields a near-zero score without any such factor.

Given the scores, we rank the $L$ blocks in descending order and retain the
Top-$K$, forming the selection set $\mathcal{K}_{n}$ whose activations are uploaded.
This reduces the payload to $K/L$ of that required for full-block transmission, while keeping
the blocks most informative for the adaptation objective and discarding the
low-importance remainder.

\begin{algorithm}[!t]
\caption{Adaptive Knowledge Distillation for Federated Low-Rank LLM over Wireless Networks}
\label{alg:main}
\begin{algorithmic}[1]
  \Require Global model $W_g$, client models $\{W_{n}\}_{n=1}^N$,
           public dataset $\mathcal{D}_p$,
           private datasets $\{\mathcal{D}_{n}\}_{n=1}^N$,
           selection criterion $c \in \{\text{token-aware},\,\text{loss-based}\}$,
           number of selected blocks $K$, number of selected samples $P$,
           number of communication rounds $T$
  \Ensure Global model $W_g$, client models $\{W_{n}\}_{n=1}^N$
  \State \textbf{Phase 1: Data Selection (executed once)}
  \If{$c = \text{token-aware}$}
    \For{each client $n = 1 \dots N$ \textbf{in parallel}}
      \State Compute token frequency $\mathbf{f}^{(n)}$ over $\mathcal{D}_{n}$ via Eq.~\eqref{eq:token_freq}
      \State Upload Top-$M$ index set $\mathcal{I}^{(n)} \gets \mathrm{TopM}(\mathbf{f}^{(n)})$ to server
    \EndFor
    \State \textbf{Server:} Aggregate $\{\mathcal{I}^{(n)}\}$ into $\tilde{\mathbf{f}}$ via Eq.~\eqref{eq:agg_freq}
    \State \textbf{Server:} $\mathcal{D}^{*}_{p} \gets \mathrm{SelectTopP}(\mathcal{D}_p, \tilde{\mathbf{f}})$
    \State \textbf{Server:} Broadcast $\mathcal{D}^{*}_{p}$ to all clients
  \EndIf
  \For{$\tau = 1 \dots T$}
    \If{$c = \text{loss-based}$}
      \State \textbf{Server:} Compute $\ell^{(\tau)}(\mathbf{x})$ for all $\mathbf{x} \in \mathcal{D}_p$ via Eq.~\eqref{eq:sample_loss}
      \State \textbf{Server:} Retain Top-$P$ samples by $\ell^{(\tau)}(\mathbf{x})$ as $\mathcal{D}^{*}_{p}$
      \State \textbf{Server:} Broadcast $\mathcal{D}^{*}_{p}$ to all clients
    \EndIf
    \State \textbf{Phase 2: Local Training and Selective Activation Upload}
    \For{each client $n = 1 \dots N$ \textbf{in parallel}}
      \State $W_{n} \gets \mathrm{LocalTrain}(W_{n}, \mathcal{D}_{n})$
      \State $\mathbf{H}_n \gets \mathrm{Infer}(W_{n}, \mathcal{D}^{*}_{p})$
      \State Compute $\mathrm{Score}_n(k)$ for all $k \in \{1,\dots,L\}$ via Eq.~\eqref{eq:block_score}
      \State $\mathcal K_n \leftarrow \mathrm{TopK}(\{\mathrm{Score}_n(k)\}_{k=1}^L)$
      \State $\widehat{\mathbf H}_n \leftarrow \{\mathbf H_n^{(k)}\}_{k\in\mathcal K_n}$
      \State Upload $\widehat{\mathbf{H}}_n$ to server
    \EndFor
    \State \textbf{Phase 3: Server Aggregation and Distillation}
    \State \textbf{Server:} $\mathcal K^* \leftarrow \bigcup_{n=1}^N \mathcal K_n$
    \State \textbf{Server:} Aggregate $\{\widehat{\mathbf{H}}_n\}$ into $\{\mathbf{H}_{\mathrm{a}}^{(k)}\}_{k \in \mathcal{K}^{*}}$ via Eq.~\eqref{eq:aggregation}
    \State \textbf{Server:} $\mathbf{H}_g \gets \mathrm{Infer}(W_g, \mathcal{D}^{*}_{p})$
    \State \textbf{Server:} Update $W_g$ by minimising $\mathcal{L}_{\text{server}}$ via Eq.~\eqref{eq:server_loss}
    \State \textbf{Phase 4: Downlink Parameter Broadcast}
    \State \textbf{Server:} Broadcast $\{\mathbf{A}_{\mathrm{g}}^{(k)}, \mathbf{B}_{\mathrm{g}}^{(k)}\}_{k \in \mathcal{K}^{*}}$ to all clients
    \For{each client $n = 1 \dots N$ \textbf{in parallel}}
      \State Overwrite local LoRA parameters at blocks $k \in \mathcal{K}^{*}$
    \EndFor
  \EndFor

\end{algorithmic}
\end{algorithm}
\setlength{\textfloatsep}{6pt}

\subsection{Public Data Selection}
\label{subsec:data_selection}

The efficiency of activation distillation depends not only on what is
transmitted but also on which public samples form the transfer set. We propose two simple and low-cost criteria for selecting useful samples from the public dataset $\mathcal{D}_p$ before distillation. The first criterion, token-aware selection, aligns the public dataset with the aggregate token distribution of the clients, determining what type of data should be covered. The second criterion, loss-based selection, ranks samples by their loss under the server model, identifying which samples are more informative for transmission. Since these two criteria capture different aspects of sample usefulness, they can be applied independently or combined. For example, loss-based ranking can be performed within the token-aligned candidate set to retain samples that are both distribution-representative and informative for distillation.

\subsubsection{Token-Aware Data Selection}
\label{subsec:token}

\paragraph{Motivation}
In federated fine-tuning, the private datasets held by different clients may have
substantially different token distributions. Such non-IID private data is the
source of client drift and aggregation bias, and it also shapes what the public
dataset must cover, since distillation transfers client knowledge only through
activations evaluated on the public samples. A fixed and uniformly sampled public
pool need not align with the token distributions the clients have adapted to, so
the activations it elicits carry little client-specific signal and yield an
unrepresentative distillation target. To reduce this mismatch, we select the
public dataset according to the collective token-level statistics of the
participating clients.

\paragraph{Token-Aware Public Data Selection}
For each client $n$, we compute a normalised token frequency vector $\mathbf{f}^{(n)} \in \mathbb{R}^{|\mathcal{V}|}$ over its local dataset $\mathcal{D}_n$ defined as

\begin{equation}
    f^{(n)}_{v} =
    \frac{\mathrm{count}(v,\, \mathcal{D}_{n})}
    {\sum_{v' \in \mathcal{V}} \mathrm{count}(v',\, \mathcal{D}_{n})},
    \label{eq:token_freq}
\end{equation}

\noindent where $|\mathcal{V}|$ is the vocabulary size. To avoid transmitting the full frequency vector, each client uploads only the index set of its Top-$M$ most frequent tokens, denoted by $\mathcal{I}^{(n)} \subset \mathcal{V}$ with $|\mathcal{I}^{(n)}| = M$. The server aggregates these index sets into a token-importance profile defined as

\begin{equation}
    \tilde{f}_{v} =
    \frac{1}{N} \sum_{n=1}^{N}
    \mathbf{1}\!\left[v \in \mathcal{I}^{(n)}\right],
    \label{eq:agg_freq}
\end{equation}

\noindent which measures how commonly token $v$ appears among the most frequent tokens of the clients. For each public sample $\mathbf{x} \in \mathcal{D}_p$, the server computes the cosine similarity between its normalised token-frequency vector and $\tilde{\mathbf{f}}$, and retains the Top-$P$ samples to form the selected public dataset $\mathcal{D}^{*}_{p}$. This procedure selects public samples that better match the shared token characteristics of the clients, while requiring only a small index-only upload from each client.

\subsubsection{Loss-Based Data Selection}
\label{subsec:loss}

\paragraph{Motivation}
Distribution alignment ensures the right mix of samples but treats all of
them as equally useful, whereas samples differ greatly in their contribution to
distillation: confidently predicted examples typically provide limited additional
corrective information, whereas higher-loss samples are more likely to expose
regions that are poorly captured by the current model~\cite{sorscher2022beyond,
mindermann2022prioritized}. We therefore rank samples according to their
predictive loss using only the public data and the server-side global model,
which requires no additional client-side computation or communication.

\paragraph{Sample Informativeness}
Let $p_g^{(\tau)}$ denote the predictive distribution of the global language model
at communication round $\tau$, with $p_g^{(0)} = p_{\mathrm{pre}}$ corresponding
to the pretrained model before federated training. For each public sample
$\mathbf{x} = (x_1, \dots, x_{T_x}) \in \mathcal{D}_p$, the server computes its
length-normalised predictive loss defined as

\begin{equation}
\ell^{(\tau)}(\mathbf{x}) = -\frac{1}{T_x} \sum_{t=1}^{T_x}
\log p_g^{(\tau)}\left(x_t \mid x_{<t}\right),
\label{eq:sample_loss}
\end{equation}

\noindent where $T_x = |\mathbf{x}|$ and the normalisation makes the score
comparable across sequences of differing length. A large
$\ell^{(\tau)}(\mathbf{x})$ indicates that the current global model explains the
sample poorly, making it a potentially informative candidate for subsequent
distillation.

\paragraph{Selection}
At each communication round, the server ranks all public samples in descending
order of $\ell^{(\tau)}(\mathbf{x})$ and retains the Top-$P$ samples as the selected
set $\mathcal{D}_{p}^{*,(\tau)}$. As the global model evolves during federated
training, the loss scores and the resulting selected subset are updated
accordingly, allowing the selected public samples to adapt to the current loss
landscape.

\subsection{Communication Analysis}
\label{subsec:comm}

The per-round uplink cost of our framework is determined by the number of
uploaded public samples $P$, the number of selected blocks $K$, and the
activation dimension $r$. Because we mean-pool over the sequence length before
transmission, the sequence axis is collapsed, so the upload is compact and
independent of the length at which the public data is processed.

The two proposed mechanisms reduce this cost along disjoint axes of the payload,
so their savings compound rather than overlap. Token-Aware Data Selection lowers
the number of transmitted samples from the full public size $|\mathcal{D}_p|$ to
$P$, and Block Importance Scoring restricts uploads to $K$ of the $L$ blocks,
giving a joint reduction factor of $(P/|\mathcal{D}_p|)\cdot(K/L)$. Either
mechanism remains effective when the other is disabled. The token index sets sent
in Phase~1 amount to only $M$ integer indices per client. Their overhead is
negligible by comparison and is incurred once per curation step rather than once
per round.

The benefit grows with model size. Unlike LoRA parameter uploads, whose volume
scales with both the hidden dimension and the number of blocks, the number of
uploaded activation vectors $P\cdot K$ is fixed by the curation and selection
budgets rather than by the architecture. Since a larger backbone has more
Transformer blocks, the retained fraction $K/L$ falls as depth increases: the
same budget that covers a sizeable share of a shallow model's blocks covers only
a small share of a deep model's, while still retaining the most informative ones.
The relative communication saving therefore grows as the model scales, which
makes the framework well-suited to deploying large language models in
bandwidth-constrained wireless settings.

\subsection{Pareto-Optimal Block-Data Allocation under a Bandwidth Budget}
\label{subsec:tradeoff}

The communication analysis above shows that the uplink volume is governed by the joint factor $P K$, where $P$ is the number of selected public samples and $K$ is the number of selected Transformer blocks. Under a fixed bandwidth budget, these two quantities are coupled: transmitting more blocks leaves room for fewer samples, and transmitting more samples requires fewer blocks. This raises a budget-allocation question that neither block selection nor data selection can answer in isolation.

To express this trade-off, we measure the activation budget in units of pooled vectors. Each pooled vector $\bar{\mathbf{h}} \in \mathbb{R}^{r}$ is encoded at $\rho$
bits per entry and therefore costs $\rho r$ bits, so the bit budget $B$ of
Eq.~\eqref{eq:comm_cost} admits

\begin{equation}
    Q = \frac{B}{\rho \, r}
    \label{eq:vector_budget}
\end{equation}

\noindent pooled vectors, and an allocation is admissible when $P \cdot K \leq Q$.

For the continuous relaxation, provided that additional public samples remain available, any unused budget can be assigned to increasing $P$, which strictly decreases the term $c_2 P^{-\gamma}$. Hence, the optimum lies on the active boundary $P K = Q$.

We model the converged cross-entropy loss as a function of the two allocation variables. Since blocks are selected in descending order of importance, we assume that the marginal task-relevant gain contributed by the $k$-th retained block decays as $a k^{-(1+\beta)}$, where $a,\beta>0$. The cumulative gain omitted when only the first $K$ blocks are retained is therefore approximated by the tail sum
$\sum_{k>K}a k^{-(1+\beta)}\approx c_1K^{-\beta}$, where $c_1=a/\beta$ follows from an integral approximation. However, retaining additional blocks may also introduce weakly informative or redundant activations into the aggregated distillation target. To first order, we model the cumulative effect of this redundancy by a linear penalty $\nu K$, where $\nu>0$ controls its overall strength, which treats the marginal noise per added block as approximately constant, consistent with the leading-order approximation used for the signal term. On the data side, samples are selected according to their relevance to the client distributions, so the additional distributional coverage provided by each newly retained sample is expected to decrease once the most representative samples have been included. These effects are captured by

\begin{equation}
    \mathcal{L}(K, P) =  \mathcal{L}_0
    + c_1 K^{-\beta}
    + \nu K
    + c_2 P^{-\gamma},
    \label{eq:loss_model}
\end{equation}

\noindent where $ \mathcal{L}_0$ is a baseline loss term independent of the block-data allocation, $c_1, c_2 > 0$ scale the contributions of blocks and samples,
$\nu$ is the per-block noise penalty, and $\beta, \gamma > 0$ control how
quickly the corresponding returns diminish.

The penalty $\nu K$ does not imply that the budget should be left partially unused. Since $c_2 P^{-\gamma}$ is strictly decreasing in $P$ and carries no counterpart penalty, any allocation with $P \cdot K < Q$ is dominated by one that spends the remainder on additional samples. The role of $\nu K$ is therefore to bound the optimal number of blocks rather than the total budget usage, and the problem is solved on the boundary $P \cdot K = Q$.

Even with the bandwidth constraint removed, Eq.~\eqref{eq:loss_model} is not
minimised by transmitting every block. The $k$-th block is worth retaining only
while its marginal gain exceeds the noise it introduces, $a k^{-(1+\beta)} >
\nu$, which places the noise-limited block count at
$ K_0 = \left( \frac{\beta c_1}{\nu} \right)^{\frac{1}{1 + \beta}}$, independent of the budget. This reproduces the behaviour reported in
Section~IV, where transmitting all blocks is inferior to a moderate
importance-ranked subset.

Substituting the active constraint $P = Q / K$ reduces the objective to a single
variable,
\begin{equation}
    \mathcal{L}(K) =  \mathcal{L}_0
    + c_1 K^{-\beta}
    + \nu K
    + c_2 \, Q^{-\gamma} K^{\gamma},
    \label{eq:loss_single}
\end{equation}

whose three budget-dependent terms represent, respectively, the signal lost to
truncation, the noise admitted by low-importance blocks, and the coverage
surrendered by the samples that each additional block displaces. Setting
$\mathrm{d}\mathcal{L}/\mathrm{d}K = 0$ balances these marginal effects and
gives the stationarity condition
\begin{equation}
    \nu K^{1 + \beta}
    + \gamma c_2 Q^{-\gamma} K^{\beta + \gamma}
    = \beta c_1,
    \qquad
    P^{\star} = \frac{Q}{K^{\star}}.
    \label{eq:optimal_split}
\end{equation}


The left-hand side of Eq.~\eqref{eq:optimal_split} increases strictly from zero without bound, so it admits a unique positive root $K_c^{\star}$. Since $\mathrm{d}\mathcal{L}/\mathrm{d}K$ changes sign from negative to positive at this point, $K_c^{\star}$ is the global minimiser of the continuous relaxation, with $P_c^{\star}=Q/K_c^{\star}$. For the discrete implementation, we evaluate each integer $K\in\{1,\dots,L\}$ with
\[
P=\min\left\{P_{\max},\left\lfloor\frac{Q}{K}\right\rfloor\right\},
\]
and select the pair with the minimum loss.



Since the second term on the left is positive, $K^{\star} < K_0$ at every finite
budget: a tight budget drives the allocation well below the noise-limited block
count, while a loosening budget raises $K^{\star}$ towards $K_0$ without
attaining it, so that beyond a moderate budget the marginal bandwidth is spent
almost entirely on samples. When the noise term is
negligible ($\nu \to 0$), Eq.~\eqref{eq:optimal_split} recovers the closed form
$K^{\star} = ( c_1 \beta / c_2 \gamma )^{1/(\beta + \gamma)}
Q^{\gamma/(\beta + \gamma)}$, in which the split adapts to the model through the
relative diminishing-return rates, a faster-saturating block term
($\beta > \gamma$) shifting the budget towards samples and a faster-saturating
data term towards blocks. The parameters $ \mathcal{L}_0, c_1, c_2, \nu, \beta, \gamma$ are fitted by
least squares to converged losses on a small calibration grid of $(K, P)$ pairs,
so that Eq.~\eqref{eq:optimal_split} accounts for the trade-off in interpretable
terms, and the bounded $K_0$ is consistent with the finding that
performance is driven by capturing the most informative blocks rather than by
their number.

The analysis so far fixes the budget and minimises loss over the split. Treating
the budget itself as a design variable turns this single optimum into a family of
efficient operating points described by two competing quantities, the
communication cost $C(K, P) = \rho r P K$, and the converged loss
$\mathcal{L}(K, P)$. Because perplexity is a monotone transform of this loss, the
dominance structure, and hence the frontier, is identical under either measure. A
configuration is dominated when another feasible configuration attains no greater
cost and no greater loss with at least one strictly smaller, and the
configurations that none dominates form the Pareto frontier \cite{sener2018multi}
{
\setlength{\abovedisplayskip}{3pt}
\setlength{\belowdisplayskip}{3pt}
\setlength{\abovedisplayshortskip}{3pt}
\setlength{\belowdisplayshortskip}{3pt}
\begin{equation} \begin{aligned} \mathcal{P} = \bigl\{ (K,P) \in &\mathcal{F} :\; \nexists\, (K',P') \in \mathcal{F} \text{ such that} \\ &C(K',P') \leq C(K,P), \\ &\mathcal{L}(K',P') \leq \mathcal{L}(K,P), \\ &\text{and at least one inequality is strict} \bigr\}. \end{aligned} \label{eq:pareto_front} \end{equation}
}\noindent where $\mathcal{F} = \{(K,P) : 1 \leq K \leq L,\; 1 \leq P \leq
P_{\max}\}$ is the grid scanned in Section~IV, defined
independently of any particular budget. At a fixed cost the constraint $PK = Q$ holds every
admissible split to the same bandwidth, so the loss minimising split of
Eq.~\eqref{eq:optimal_split} is the only non-dominated allocation at that cost,
while every other split, carrying equal cost at higher loss, lies in the
interior. Dominance by a cheaper configuration is excluded because the optimised loss
decreases strictly in the budget: the envelope theorem gives
$\mathrm{d}\mathcal{L}(K^{\star}(Q))/\mathrm{d}Q = -\gamma c_2 Q^{-\gamma-1}
(K^{\star})^{\gamma} < 0$.

Sweeping the budget therefore traces $\mathcal{P}$ through the family
$\{(\, \rho r Q,\; \mathcal{L}(K^{\star}(Q)) \,)\}_{Q}$, along which $K^{\star}$
increases towards $K_0$ without attaining it, so that at large budgets the
frontier is traced by the sample budget alone. The stationarity condition
defining the optimal split therefore coincides with the condition for an
allocation to lie on the frontier.

\section{ Numerical Results }

In this section, we evaluate the effectiveness of our proposed FLoKD framework and compare it with several existing baseline approaches.
In the simulations, we utilize the GPT-2 \cite{radford2019language} series model as the primary architecture for language modeling. To validate the effectiveness of the proposed algorithm, we use Wikitext-103~\cite{merity2016pointer}, PTB~\cite{mikolov2010recurrent}, and dialogue dataset DailyDialog~\cite{li2017dailydialog}. Wikitext-103 is a large-scale language modelling benchmark containing over 100 million tokens extracted from Wikipedia articles, while PTB is a compact and widely used language modelling benchmark derived from the Penn Treebank corpus, containing approximately one million words from Wall Street Journal text. The dialogue dataset is included to evaluate the proposed method under conversational data distributions, which exhibit different token patterns and stronger distributional variation from conventional language modelling benchmarks.

The experimental setup consists of a pool of 30 clients and a single central server, where each client hosts a local GPT-2 model. The training data is evenly partitioned among the 30 clients under a Non-IID setting, reflecting the heterogeneous data distributions commonly encountered in real-world federated learning scenarios. Additionally, a shared public dataset consisting of 2,000 samples is made available to all clients for knowledge distillation. In each communication round, a random subset of N = 10 clients is selected to participate in local model training and aggregation. Other key simulation parameters are listed in Table \ref{t1}. We evaluate the converged server perplexity (PPL) on the WikiText, PTB and Dialog dataset under varying bandwidth constraints, where lower PPL indicates better language modeling performance.

\begin{table}[htbp]
  \centering
  \caption{Experimental Settings}
  \label{t1}
  \renewcommand{\arraystretch}{1.15}
  \setlength{\tabcolsep}{4pt}
  \scriptsize
  \begin{tabular}{@{}llll@{}}
    \toprule
    \textbf{Parameter} & \textbf{Value} & \textbf{Parameter} & \textbf{Value} \\
    \midrule
    Datasets        & \multicolumn{3}{l}{WikiText-103, PTB, Dialog} \\
    Backbone        & \multicolumn{3}{l}{GPT-2 (12 Transformer blocks)} \\
    Evaluation      & \multicolumn{3}{l}{Perplexity (PPL), 3 runs averaged} \\
    \midrule
    LoRA rank ($r$) & 8     & Client pool size     & 30 \\
    LoRA $\alpha$   & 32    & Participating clients ($N$) & 10 \\
    LoRA dropout    & 0.1   & Comm.\ rounds      & 50/100 \\
    Learning rate   & 0.001 & Blocks sent ($K$)  & 3/2 \\
    Batch size      & 64    & Weight ($\lambda$) & 0.03 \\
    Seq.\ length    & 512   & Local samples      & 2\,000 \\
    Weight decay    & 0.001 & Public samples     & 1\,000 \\
    \bottomrule
  \end{tabular}
  \vspace{-5pt}
\end{table}

\textbf{Baselines} To comprehensively evaluate FLoKD under bandwidth-constrained federated fine-tuning for generative large language models, we compare it against three representative baselines that differ in what type of information is transmitted during collaboration, namely output-level knowledge, partial LoRA parameters, and full LoRA parameters. This setup enables a systematic comparison of the communication--performance trade-off across different transmission paradigms.

\begin{itemize}
    \item \textbf{FedMKT}~\cite{fan2025fedmkt}: FedMKT is a federated mutual knowledge transfer baseline that exchanges logits instead of model parameters. In each communication round, clients fine-tune locally and generate prediction logits on shared public samples. To reduce communication overhead, only the Top-$K$ logits are transmitted to the server, where they are aggregated to guide collaborative knowledge transfer. In our experiments, FedMKT serves as a representative output-level baseline for evaluating logit-based federated distillation under limited bandwidth.

    \item \textbf{FedPET}~\cite{zhang2023fedpetuning}:
    FedPET is used as the standard full LoRA parameter aggregation baseline. Each client performs local parameter-efficient fine-tuning by updating its LoRA adapters on private data, and all trainable LoRA parameters are uploaded to the server after local training. The server aggregates the received LoRA updates to form global adapter weights, which are then broadcast back to clients for the next round. Compared with full-model federated learning, FedPET reduces communication by restricting transmission to LoRA modules, while still following the conventional parameter exchange paradigm.

    \item \textbf{FedSA}~\cite{guo2025selective}:
    FedSA adopts a selective aggregation strategy for LoRA-based federated fine-tuning. Specifically, only the LoRA A matrix is transmitted to and aggregated at the server, while the LoRA B matrix remains local on each client. This design reduces the communication payload compared with full LoRA parameter exchange and serves as a representative partial-parameter transmission baseline. In this work, FedSA is used to evaluate whether communicating only a subset of LoRA parameters can achieve a favorable balance between communication efficiency and model performance.

\end{itemize}

Unless otherwise specified, all baselines are implemented under the same training protocol, model backbone, data partition, local update steps, and communication schedule as FLoKD. This unified experimental setting ensures a fair comparison and allows us to precisely assess whether FLoKD can achieve a better trade-off between communication efficiency and downstream generation performance than existing logit-based and LoRA-based federated fine-tuning methods.


\begin{figure}[htbp]
\centering
\includegraphics[width=\columnwidth]{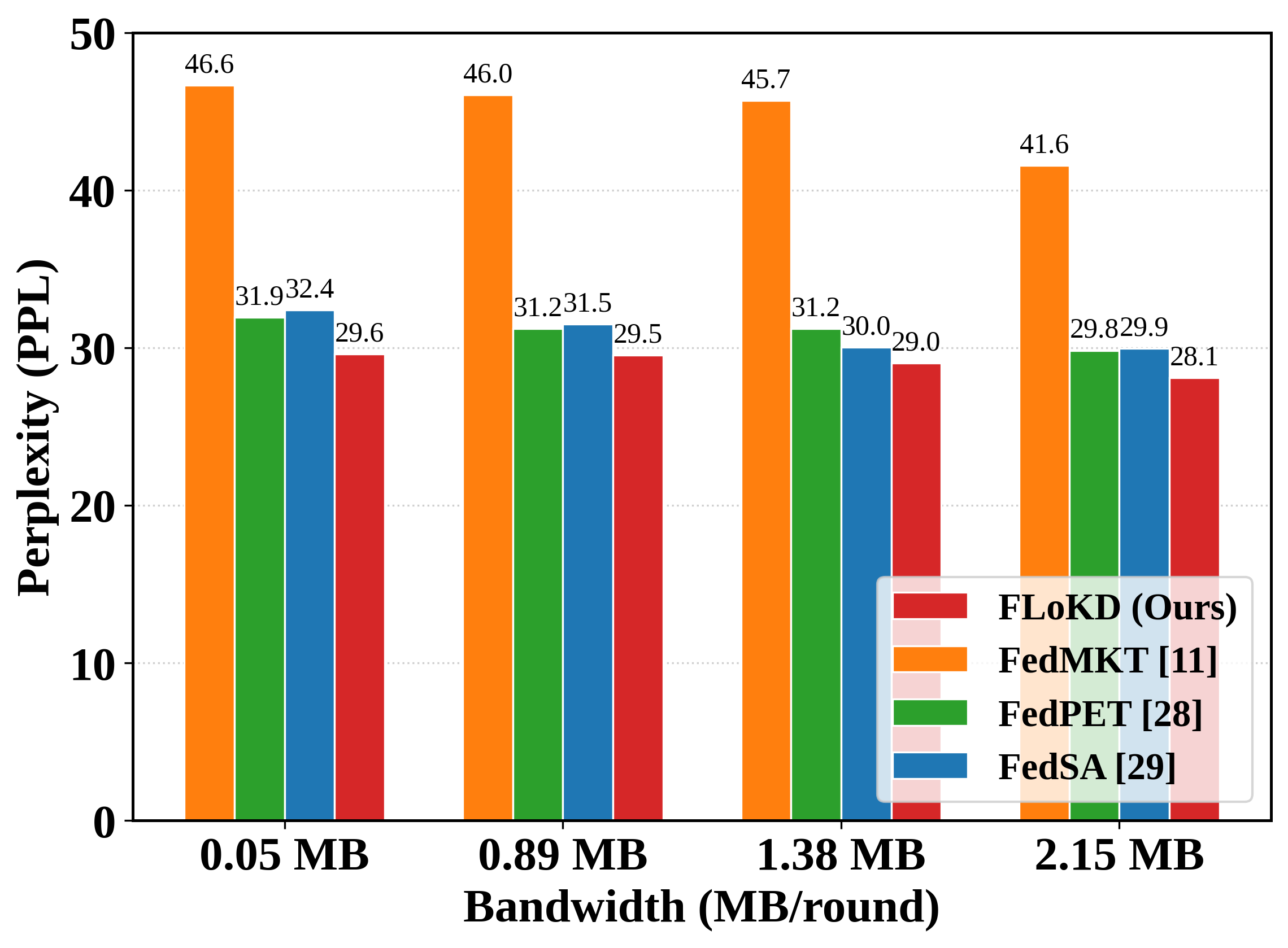}
\caption{Performance of FLoKD and Baseline frameworks under Limited Bandwidth on WikiText-103}
\label{fig2}
\end{figure}

We evaluate the per-round upload budget at four settings: $\{0.05, 0.89, 1.38, 2.15\}$~MB. The two intermediate budgets, 0.89~MB and 1.38~MB, approximately cover the native communication requirements of the compared methods, including full LoRA parameter upload and activation-based transmission, while the extreme settings of 0.05~MB and 2.15~MB probe behaviour under severely constrained and near-unconstrained uplinks. Since the compared methods transmit different quantities, a common budget is enforced by truncating each method's native payload so that all methods occupy an identical uplink volume per round. For FedSA and FedPET, which upload LoRA parameters, the entries with the largest magnitude are retained until the budget is met, and the remaining entries are left unchanged at the server for that round. For FedMKT, which uploads logits over the public samples, only the Top-$k$ logits of each token are retained, with $k$ determined by the budget. For the proposed framework, the budget is met through the joint block and data selection of Section~III, which allocates the available volume between the transmitted blocks $K$ and the selected samples $P$. The differences reported in Fig.~\ref{fig2} therefore reflect how effectively each method allocates a fixed uplink volume rather than differences in the amount of information transmitted.

As illustrated in Fig.~\ref{fig2}, the four methods exhibit markedly different sensitivities to bandwidth limitations. FedMKT consistently yields the highest PPL across all bandwidth settings. Since the vocabulary of LLMs is typically very large, the resulting logits are correspondingly high-dimensional, so constrained bandwidth forces a large portion to be discarded, transmitting only coarse and incomplete output-level supervision, which worsens as the budget shrinks. In contrast, the FedPET baseline, which transmits LoRA parameters, maintains relatively stable performance across bandwidth settings but remains consistently inferior to our method, suggesting that transmitting parameter updates without accounting for their informational significance yields limited accuracy. FedSA, which transmits only the LoRA down-projection matrices, achieves competitive performance under moderate bandwidth but degrades noticeably when the budget becomes extremely limited, indicating that substantial information is lost when parameters are aggressively compressed. Our proposed method, FLoKD, achieves substantially lower PPL than all baselines across the entire bandwidth spectrum and, critically, exhibits remarkable robustness to bandwidth reduction, as the performance gap between low and high bandwidth settings remains consistently narrow. This behavior indicates that FLoKD effectively prioritizes the transmission of the most informative components, enabling the global model to converge to a high-quality solution even under severely constrained communication budgets.

\begin{figure}[htbp]
\centering
\includegraphics[width=\columnwidth]{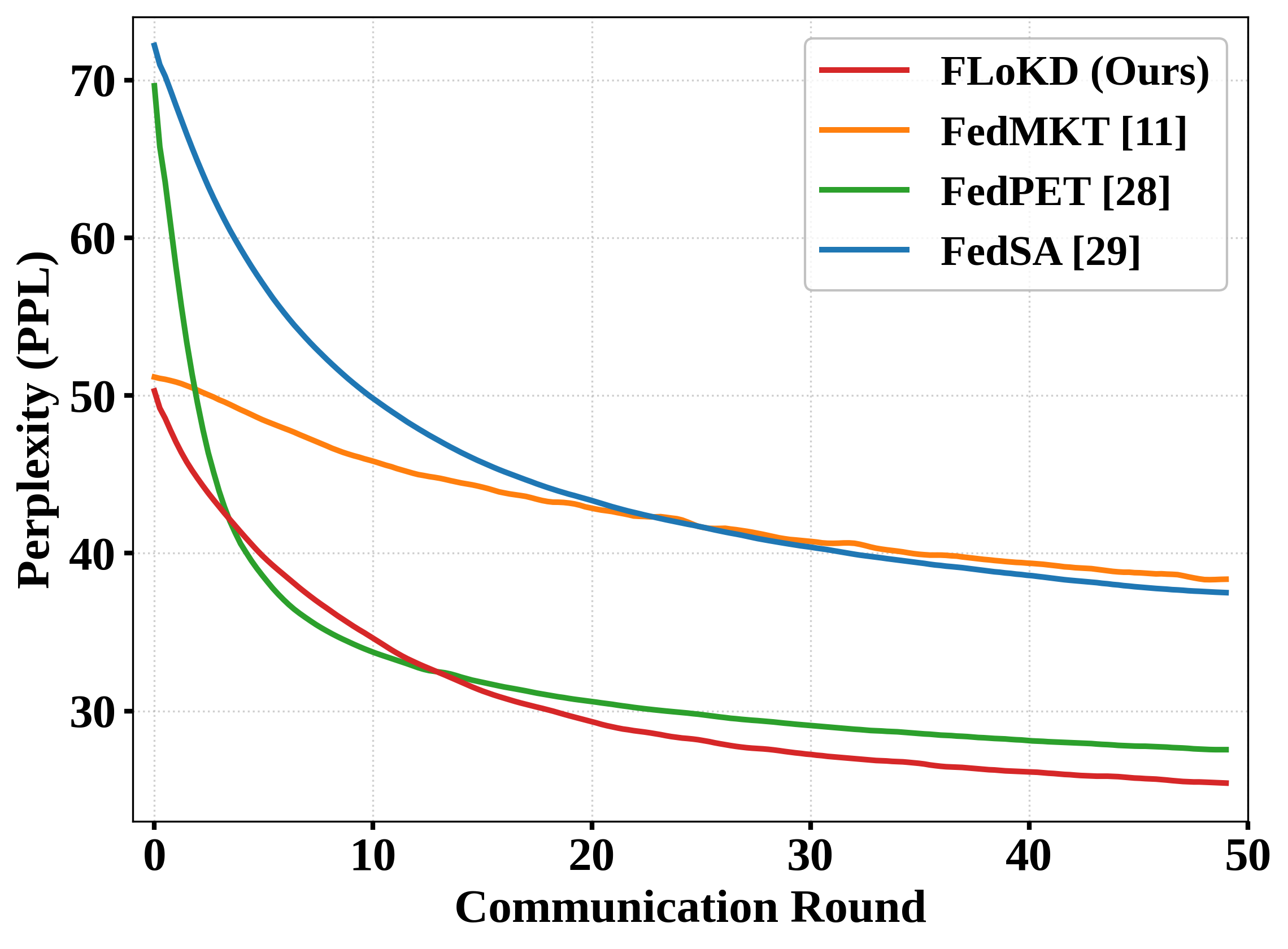}
\caption{ Convergence Behavior of FLoKD Against Three Baseline Methods }
\label{fig3}
\end{figure}

Fig.~\ref{fig3} illustrates the convergence of the server-side perplexity (PPL) under the unlimited-bandwidth setting, where the proposed FLoKD is compared with three baselines. FLoKD exhibits a decrease in PPL during the first ten communication rounds, remaining comparable to the strongest baseline throughout this early phase, and continues to decrease gradually before settling at the lowest final PPL of approximately 25. FedPET is the most competitive baseline and descends at a similar rate in the early rounds, although it stabilizes at a slightly higher PPL of around 27. Consequently, FLoKD reaches a given perplexity level within fewer communication rounds and thus at a lower communication cost. In contrast, FedMKT and FedSA reduce PPL considerably more gradually after the initial rounds and plateau at a markedly higher level of around 38. These results indicate that distilling intermediate LoRA activations yields both a rapid early reduction and a lower, more stable final perplexity than the baseline methods, even when communication is unconstrained.

\begin{figure}[htbp]
\centering
\includegraphics[width=\columnwidth]{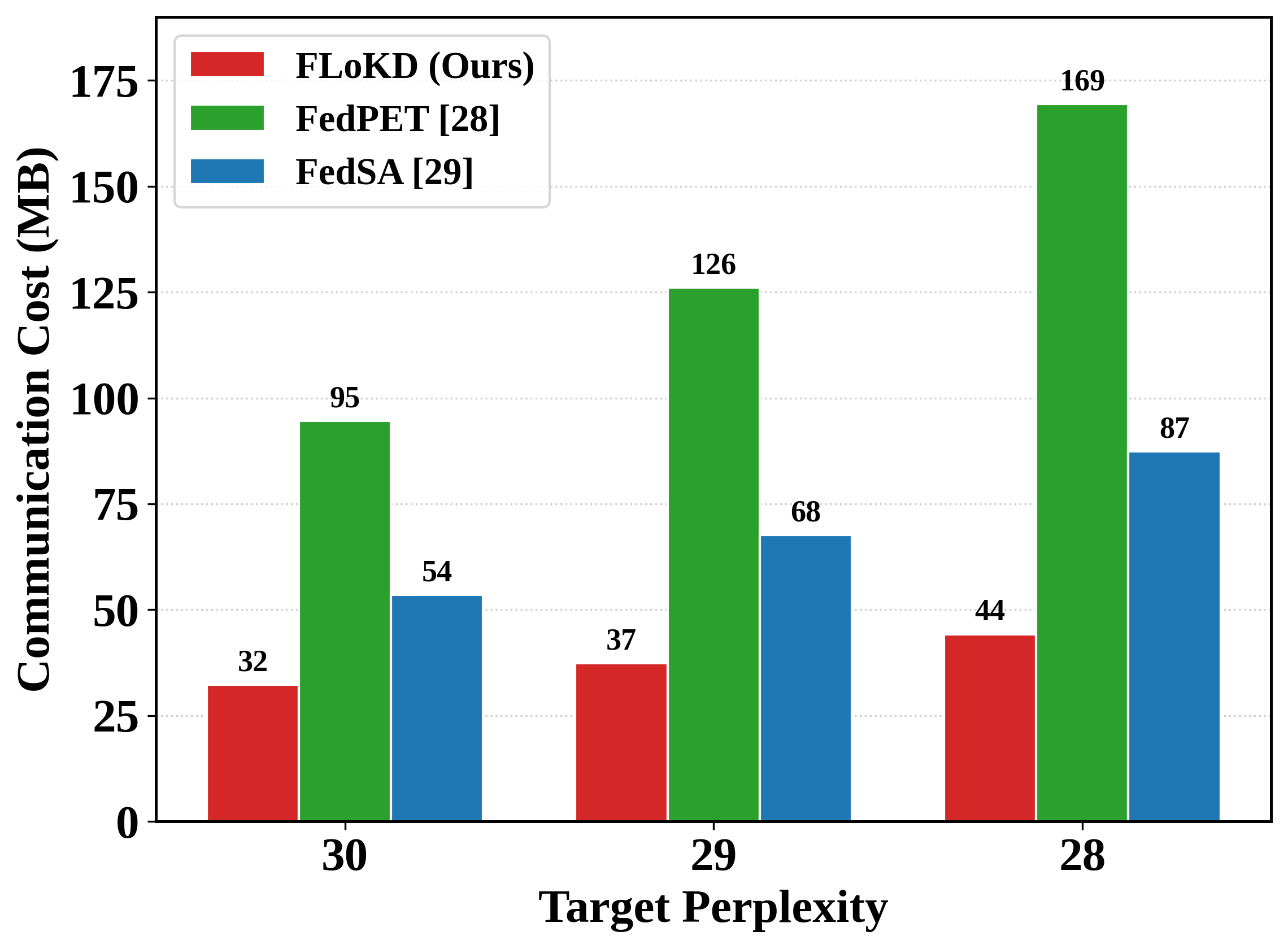}
\caption{Communication Cost for Target PPL (lower is better)}
\label{fig4}
\end{figure}

Fig.~\ref{fig4} compares the communication cost required to reach different target perplexity levels. FLoKD consistently achieves the lowest communication overhead across all target PPL settings, showing that it can reach the same generation quality with substantially less transmitted information. Compared with FedPET, FLoKD reduces the required communication cost by more than 50\% across all three target perplexity levels, demonstrating that the proposed adaptive knowledge distillation strategy can effectively remove redundant transmissions while preserving the most informative components. As the target perplexity becomes more stringent, all methods require higher communication budgets, but FLoKD maintains a clear advantage, confirming its superior communication efficiency under bandwidth-constrained federated learning.

\begin{figure}[htbp]
\centering
\includegraphics[width=\columnwidth]{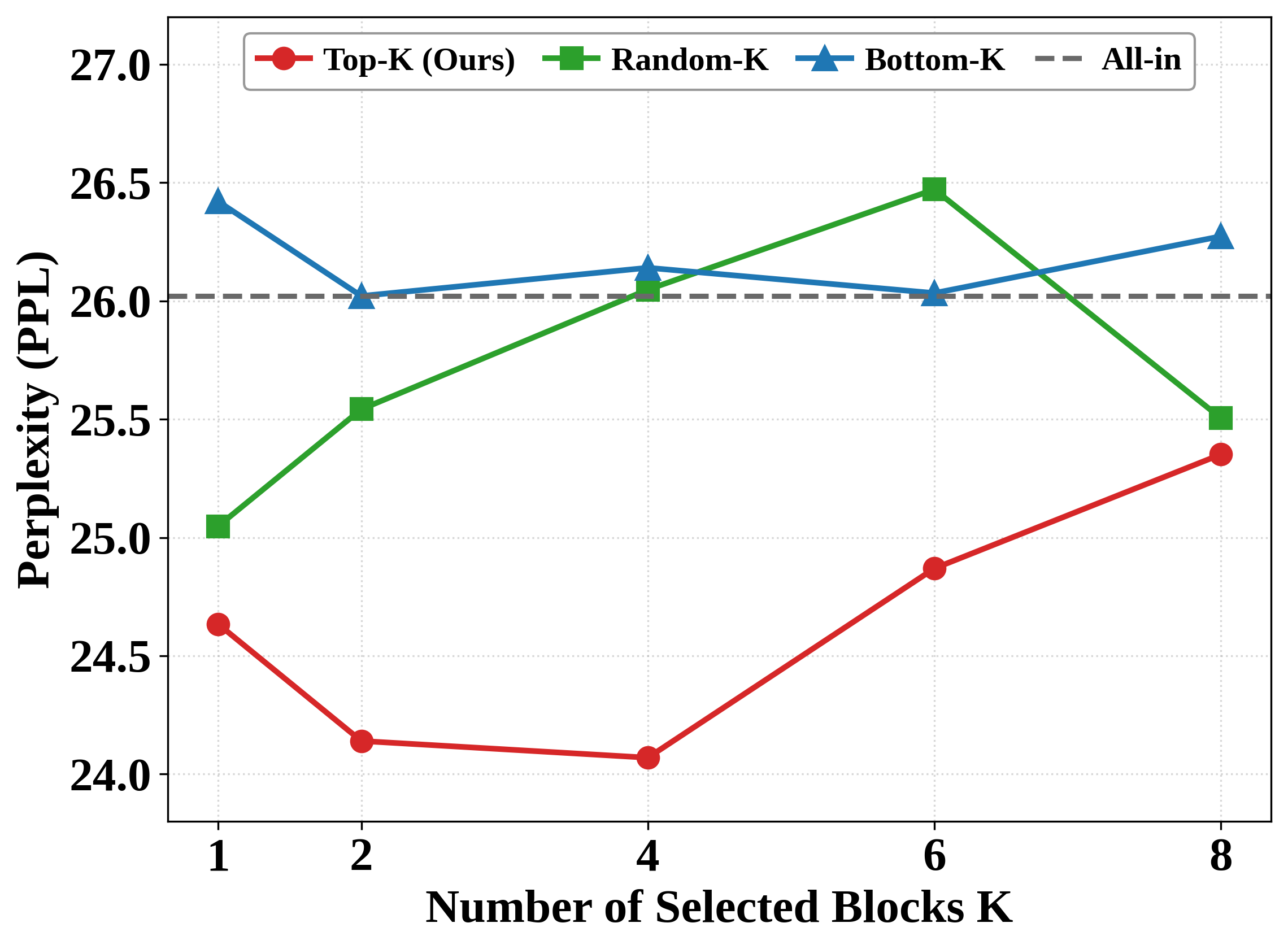}
\caption{Performance Comparison of the Proposed Block Selection and Other frameworks }
\label{fig5}
\end{figure}

To evaluate the effectiveness of the proposed block importance scoring framework in identifying informative Transformer blocks, we compare the server PPL under different block selection strategies, including Top-K, Random-K, and Bottom (Bottom-K) selection. Fig.~\ref{fig5} presents the effect of these strategies. Top-K consistently achieves lower server PPL than Random-K and Bottom-K across different numbers of selected blocks, indicating that the proposed contribution score can effectively rank Transformer blocks according to their utility for activation-based distillation. In contrast, Bottom-K generally leads to the worst performance, which further confirms that low-contribution blocks provide limited useful information to the global model.

Moreover, the best performance is obtained when selecting a moderate number of blocks, while further increasing K does not continuously improve performance. Since this experiment is conducted under an unconstrained bandwidth setting, this suggests that additional blocks may bring diminishing marginal information gains and may introduce redundant or less informative representations. Overall, the results validate the effectiveness of contribution-based block selection and show that selecting a compact set of high-contribution blocks is sufficient for effective distillation.

\begin{figure}[htbp]
\centering
\includegraphics[width=\columnwidth]{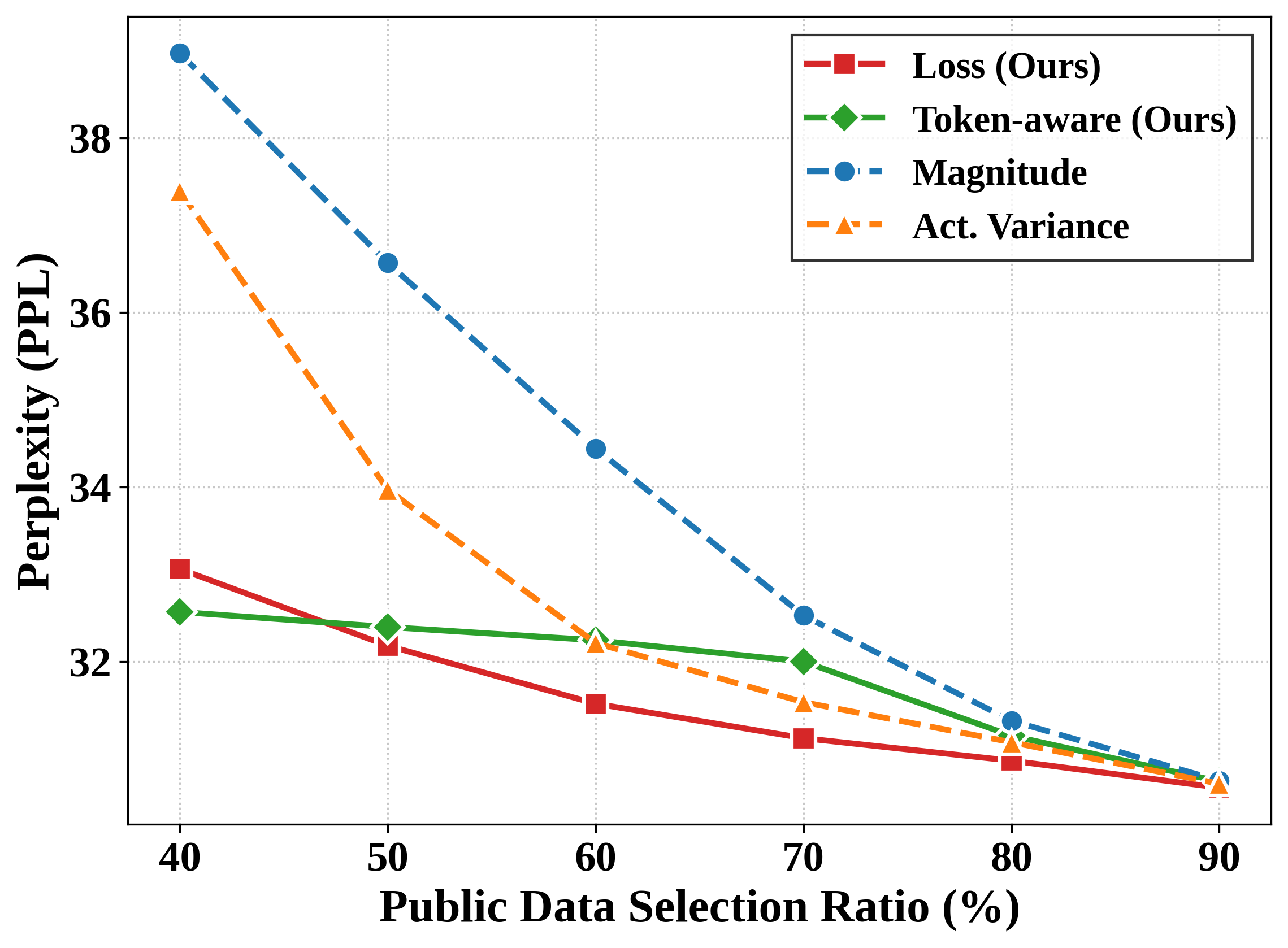}
\caption{
Comparison of Data Selection Methods under Different Selection Ratios (with Mixed PTB \& DailyDialog Dataset)}
\label{fig6}
\end{figure}

Fig.~\ref{fig6} compares different public data selection strategies for federated LLM learning under a more heterogeneous non-IID setting, where the public candidate pool is constructed by mixing PTB with dialogue-domain data. The results show that public data selection clearly affects global model performance, especially when the selection rate is low. Magnitude-based selection performs poorly under aggressive data reduction, with PPL remaining high at 38.97 and 36.56 under 40\% and 50\% selection rates, respectively. This indicates that activation magnitude alone is not a reliable criterion for identifying useful public samples. Activation-variance selection improves when more data are selected, but its performance is unstable in the low-selection regime. In contrast, loss-based selection achieves strong performance across different selection rates and obtains one of the lowest PPL values at high selection rates, reaching 30.55 at 90\%. The token-aware strategy is also highly competitive and is particularly effective when the public data budget is limited, achieving the lowest PPL of 32.57 at a 40\% selection rate. This suggests that token-level distributional information provides a useful signal for preserving representative public samples under strict data constraints. Overall, loss-based selection provides a reliable general-purpose criterion for public data selection, while token-aware selection is particularly useful under aggressive selection budgets or one-shot curation, where token-level statistics offer a low-cost signal for retaining representative samples.

\begin{figure}[htbp]
\centering
\includegraphics[width=\columnwidth]{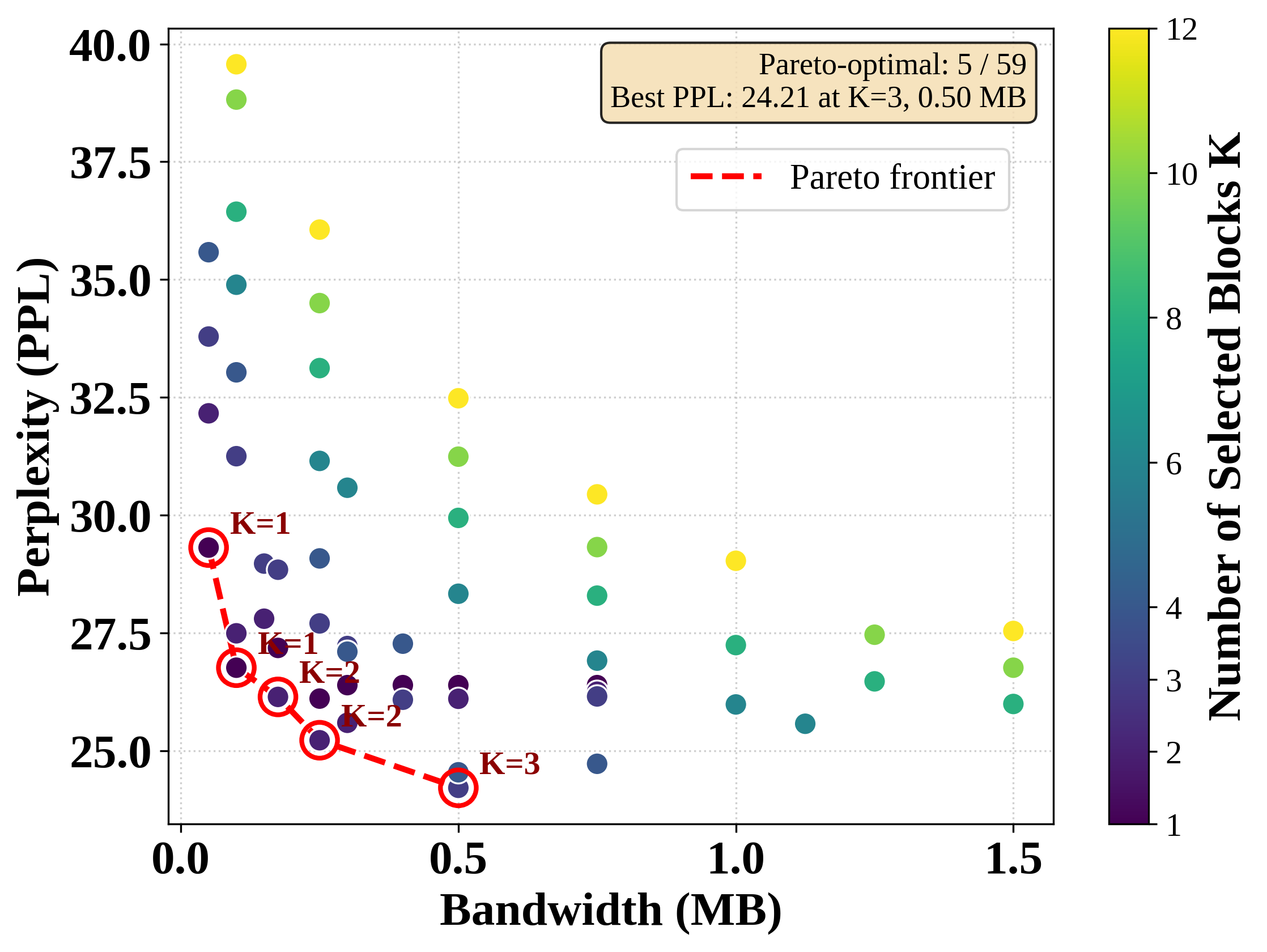}
\caption{Pareto Frontier: Bandwidth vs Perplexity (59 tests) }
\label{fig7}
\end{figure}

As shown in Fig.~\ref{fig7}, we conduct a Pareto analysis to examine the trade-off between selected public dataset coverage and the number of transmitted blocks in federated activation distillation, evaluated by communication bandwidth and final PPL. Since communication cost depends on both selected public data and transmitted blocks $K$, the bandwidth must be split between public data coverage and block-wise activation transmission. It should be noted that the public dataset is finite and does not increase indefinitely. For example, in our setting, the $K=1$ configuration saturates at around 0.50 MB, corresponding to transmitting activations from one Transformer block over the full available public dataset. The Pareto frontier reveals a clear allocation pattern across different bandwidth regimes. Under the most restrictive budgets, it is more effective to allocate bandwidth to broader public dataset coverage rather than to activations from additional Transformer blocks. Accordingly, $K=1$ gives the optimal configurations, achieving a PPL of 29.31 at 0.05 MB and 26.76 at 0.10 MB. When more bandwidth is available, the optimal setting shifts to $K=2$, reducing the PPL to 26.14 at 0.175 MB and 25.22 at 0.25 MB. The best observed result is obtained with $K=3$, which reaches a PPL of 24.21 at 0.50 MB. Beyond this range, increasing the number of transmitted blocks no longer improves the trade-off. The best $K=4$ configuration reaches only 24.72 PPL at 0.75 MB, while $K=12$ requires 1.50 MB but still gives a worse PPL of 27.54. These dominated configurations indicate that full-block activation transmission is inefficient. A compact subset of informative Transformer blocks is sufficient for effective distillation, whereas excessive block transmission consumes bandwidth that is better used to preserve useful public dataset coverage under constrained communication budgets.

\section{Conclusion}
This paper proposed an adaptive knowledge distillation framework for federated low-rank LLM fine-tuning over wireless networks. Instead of transmitting full model parameters or output logits, the framework uses intermediate LoRA activations as the distillation signal, enabling effective knowledge transfer with lower communication overhead. To further improve efficiency, we introduced a Transformer block importance scoring framework to transmit only informative block activations, together with a token-aware data selection strategy to select the shared public dataset. Extensive experiments on language generation tasks across multiple datasets show that the proposed framework achieves fast convergence and better PPL while reducing communication overhead by 50--65\% compared with baselines. These results demonstrate the effectiveness of communication-aware activation distillation for bandwidth-constrained federated LLM fine-tuning.

\IEEEpubidadjcol

\bibliographystyle{IEEEtran}
\bibliography{IEEEabrv,Bibliography}

\newpage

 


\vfill

\end{document}